# ColdRoute: Effective Routing of Cold Questions in Stack Exchange Sites

Jiankai Sun · Abhinav Vishnu · Aniket Chakrabarti · Charles Siegel · Srinivasan Parthasarathy

**Abstract** Routing questions in Community Question Answer services (CQAs) such as Stack Exchange sites is a well-studied problem. Yet, cold-start – a phenomena observed when a new question is posted is not well addressed by existing approaches. Additionally, cold questions posted by new askers present significant challenges to state-of-the-art approaches. We propose ColdRoute to address these challenges. ColdRoute is able to handle the task of routing cold questions posted by new or existing askers to matching experts. Specifically, we use Factorization Machines on the one-hot encoding of critical features such as question tags and compare our approach to well-studied techniques such as CQARank and semantic matching (LDA, BoW, and Doc2Vec). Using data from eight stack exchange sites, we are able to improve upon the routing metrics (Precision@1, Accuracy, MRR) over the state-of-the-art models such as semantic matching by 159.5%,31.84%, and 40.36% for cold questions posted by existing askers, and 123.1%, 27.03%, and 34.81% for cold questions posted by new askers respectively.

**Keywords** question routing · expert finding · cold-start problem · question answering services

# 1 Introduction

Nowadays, the Community-based question answering sites (CQAs) such as Stack Overflow, Stack Exchange Sites, and Quora, which enable people to post questions and answers in various domains [Yang et al., 2013] have accumulated millions

Jiankai Sun, Srinivasan Parthasarathy
The Ohio State University
E-mail: sun.1306@osu.edu, srini@cse.ohio-state.edu

Abhinav Vishnu, Charles Siegel
Pacific Northwest National Laboratory
E-mail: abhinav.vishnu@pnl.gov, charles.m.siegel@gmail.com

Aniket Chakrabarti
Microsoft (work done while at The Ohio State University)
E-mail: chakrabarti.14@osu.edu

of questions and posted answers over time [Zhao et al., 2016, Zhao et al., 2017, Song et al., 2017]. One important task in CQAs is to make recommendations for new questions (routing questions), that fall in three scenarios : 1) find experts. 2) route questions to the right answers (identification of best answers). 3) find similar questions to new questions [Yang et al., 2013]. In this paper, we focus on the problem of expert finding [Xu et al., 2012, Zhao et al., 2013, Yang et al., 2013, Fang et al., 2016, Zhao et al., 2016, Zhao et al., 2017], which is to choose the right experts for answering questions posted by users in Stack Exchange, which is a network of question-and-answer (Q&A) websites containing topics in various fields. Each Stack Exchange site covers a specific topic. For example, site Physics [1] accumulates all questions about physics.

Usually there are two types of questions in CQAs – resolved (questions with answers) and newly posted questions (questions that have not received any answers). The newly posted questions may themselves be posted by new askers (such as new registered users who have not asked a question earlier) or existing askers (such as users who have asked several questions previously). We refer to these kinds of questions as *cold questions*. The majority of approaches have focused on evaluating content quality after the fact (after questions have been resolved) [Yang et al., 2013]. Yet, as the Stack Exchange sites continue to grow, routing the cold questions to matching experts before answers have been provided has become a critical problem. We refer to this problem as a cold start problem, which is also a common problem in recommender systems [Sun et al., 2012, Wang et al., 2012, Wang et al., 2014b, Cheng et al., 2017].

1.1 Related Approaches: Semantic Matching

One possibility to handle a cold question is to consider its textual information. This has already been proposed with *semantic matching* (SM), which falls into two categories [Srba and Bielikova, 2016]: language model-based [Li and King, 2010, Li et al., 2011, Dong et al., 2015], and topic model-based [Yang and Manandhar, 2014, Szpektor et al., 2013, Yang et al., 2013] question routing.

SM can rank the answerers for a given question based on their semantic relevance (i.e. cosine similarity). Questions and answerers (based on all answers or best answers posted by the user) are represented by semantic models such as Bag of Words (BoW) [Figueroa and Neumann, 2013, Zhou et al., 2013], Latent Dirichlet Allocation (LDA) [Guo et al., 2008, Ji et al., 2012], Word2Vec [Mikolov et al., 2013], and Doc2Vec [Le and Mikolov, 2014] [2]. These matching models have demonstrated their power on finding suitable experts recently [Srba and Bielikova, 2016]. However, on average only a few users show their opinions for each question in CQAs and it is costly to construct a sparse user-question matrix for latent topic models such as LDA [Liu et al., 2017]. Although SM models can address the issue of the lexical gap between the user profiles and posted question, it is undeniable that they fail to overcome the sparsity of CQAs data [Liu et al., 2017].

This conclusion is consistent with our experiments in Stack Exchange sites as demonstrated in Figure 1, which shows the $Precision@1$ performance by selecting the answer which has the highest semantic relevance score as the best answer on eight different Stack Exchange sites [3]. We use BoW, LDA, and Doc2Vec in our experiments to represent questions and answers and compute relevance scores [4]. The evaluation measure is $Precision@1$, which computes the average number of times that the best answer (answerer) is ranked in top-1 by a certain semantic-matching based model (please refer to equation 12 for more details). In Figure 1 we observe that the best $Precision@1$ of semantic matching is less than 30%. This indicates that leveraging textual information solely plays a limited role in the identification of best answers *(answerers)* in CQAs.

---

[1] https://physics.stackexchange.com/
[2] More technical details can be viewed at Section 5.3

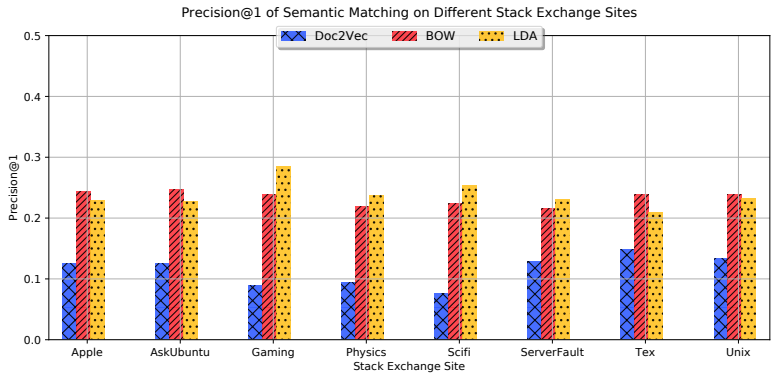

Fig. 1: Precision@1 of semantic matching based models (BoW, LDA, Doc2Vec) on different Stack Exchange sites. The best $Precision@1$ of semantic matching is less than 30%.

1.2 Voting score as the metric of finding experts

In question routing, we need to identify the metric of finding the best answerer. One possibility is by using the number of up-votes and down-votes. In Stack Exchange sites, voting is central for providing quality questions and answers [5]. Voting up a post signals to the rest of the community that the post is interesting, well-researched, and useful. A highly voted post reflects the quality of the post – which may be viewed by the future visitors. The more that people vote on a post, the more certain future visitors can be confident of the quality of information contained within the post. Hence voting indicates a CQA community's long-term review for a given user's expertise level under a specific topic. Users with high expertise tend to receive high votes for their Q&A posts [Anderson and et al., 2012, Yang et al., 2013]. Each voting score is an integer, which is calculated based on the difference between corresponding answer's up-votes and down-votes which are assigned to it by users who viewed the question or provided answers in the CQAs.

---

[3] Other Stack Exchange sites demonstrate a similar trend. To reduce space usage, we report eight large and popular Stack Exchange sites in our paper.

[4] As Doc2Vec is heavily related to Word2Vec, we only reported Doc2Vec in our experiments

[5] https://stackoverflow.com/help/why-vote

In Stack Exchange sites, askers can select a solution as the best answer for their asked questions. The user who provided the best answer is represented as the best answerer. We conducted experiments to analyze the correlation between answers' voting score and whether they are selected as the best answers in Stack Exchange sites.

Given a question $q$ [6], UpVotes-Rank selects the answerer who has the highest voting score in $q$'s answering thread as $q$'s best answerer. We then use Precision@$k$ to measure the average number of times that the best answerer is ranked in top-$k$ in terms of voting scores, where $k = 1, 2, 3$. In Figure 2 we can see that about 70% best answerers are in top 1 ranked by voting score, about 85% best answerers are in top 2, and 95% are in top 3. Hence, it indicates that we can view the problem of identification of best answerers as finding the answerers who have the highest predicted voting scores.

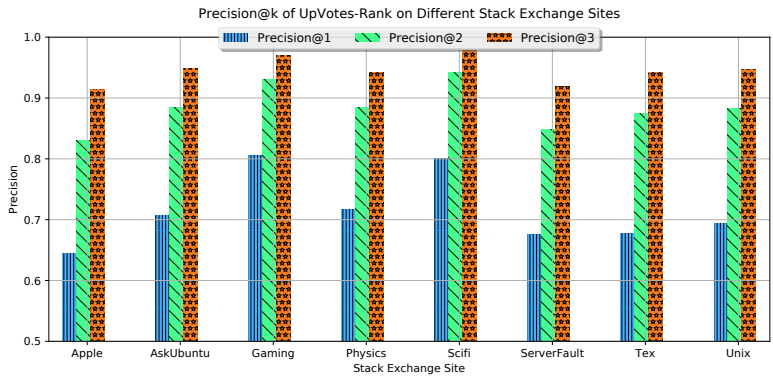

Fig. 2: *Precision*@$k$ ($k = 1, 2, 3$) of UpVotes-Rank on different Stack Exchange sites. About 70% best answers are in top 1 ranked by number of voting score, about 85% best answers are in top 2, and about 95% are in top 3.

Up to now we have concluded that voting score modeling is a highly feasible approach. Several state-of-the-art approaches learn their question routing models by using received number of up-votes and down-votes of their past question-answering activities [Yang et al., 2013, Zhao et al., 2015b, Zhao et al., 2016, Zhao et al., 2017]. However, these approaches are not easily transferable to do expert finding for cold questions, which will be discussed in Section 2.

1.3 Problem Definition

Our evaluation has indicated that a simple application of the proposed solutions (semantic matching based models) to cold question is ineffective. So what are the approaches for doing voting score modeling in the absence of an answer? *This leads us to the following challenges: 1) What are the features that determine the routing*

---

[6] Questions which have at least five answers are selected for evaluation

*of cold questions?, 2) Which algorithms are potentially effective for routing cold questions using these features?*

1.4 Overview of ColdRoute

In this paper, we undertake these challenges. We propose ColdRoute – a framework that combines cold questions' limited information (askers, questions tags, and textual descriptions) in a unified framework and leverage Factorization Machines (FMs) to address the sparsity evident in these features. As shown in the Figure 1, textual description plays a limited role in semantic matching models. We leverage question tags, rather than textual descriptions, in our model. A tag is a word or phrase that describes the topic of the question in CQAs[7]. Hence tags are important user-generated category information that achieves fine-grained and dynamic topic representation. Users who use a particular tag when posting questions or answers might prefer topic summaries most relevant to that tag [Ramage et al., 2009]. Incorporating tags of questions and answers into textual content aids in better discovery of user topical interests [Yang et al., 2013].

Each answering thread between a question and an answerer can be represented as a quadruple of the target question, its asker, the corresponding answerer and question tags. A simple approach is to encode these answering threads using one-hot encoding. However, one-hot encoding can cause sparsity problem, which is not handled effectively by several Machine Learning algorithms. Rendle et al. [Rendle, 2012, Rendle, 2010] proposed to use FMs to handle data sparsity problems in recommender systems. While applying FMs for cold questions routing has not been well studied, we propose ColdRoute which is based on FMs to model all possible interactions between variables (questions, askers, answerers, and question tags) in sparse quadruples. Extensive experiments on Stack Exchange sites demonstrate the improved efficacy of our approach over contemporary state-of-the-art models in the tasks of question routing and identification of best answerers for newly posted questions no matter whether they are asked by new askers or existing askers.

1.5 Our Contributions

Specifically, we make the following contributions in the paper:

- We present a simple feature encoding which requires readily available information such as question tags, asker's information, question title and body.
- Our simple encoding introduces sparsity. Hence, we consider a set of machine learning approaches and leverage FMs, since they address the sparsity problem effectively. FMs are also able to model all interactions from users' past activities in sparse settings.
- We iteratively introduce features and present their relative importance in cold question routing.

---

[7] https://stackoverflow.com/help/tagging

– We compare our approach with MC [Zhao et al., 2015b] which uses social network information. We observe that ColdRoute outperforms MC, which makes ColdRoute amenable for practical deployments, since social network information is typically difficult to access.

Our extensive experiments indicate that our model can improve upon the routing metrics (Precision@1, Accuracy, MRR) over the state-of-the-art models such as semantic matching by 159.5%, 31.84%, and 40.36% for cold questions posted by existing askers, and 123.1%, 27.03%, and 34.81% for cold questions posted by new askers respectively. We observe that tags are critical in cold routing question, and surprisingly more effective than FMs on question's title and body itself.

## 2 Related Work

In this section, we present related work for ColdRoute. Existing work can be divided into two groups for user expertise estimation: the *authority-oriented* approaches, and the *topic-oriented* approaches [Zhao et al., 2015b].

The authority-oriented user expertise estimation methods are based on link analysis for the ask-answer relation between users [Zhang et al., 2007, Yang et al., 2008, Zhu et al., 2011, Sung et al., 2013, Zhu et al., 2014]. Zhang et al. built a graph based on asker-answerer relationships for a set of threads in Java Developer Forum[8] and leveraged several network-based ranking algorithms like PageRank, HITs, InDegree, etc. to discover users' expertise [Zhang et al., 2007]. Yang et al. proposed to construct a prestige graph of tasks and users [Yang et al., 2008]. Each user's relative expertise would be determined by the standard PageRank algorithm. Kumar et al. [Kumar and Pedanekar, 2016] created a directed graph of asker-answerer pairs and then leveraged the PageRank algorithm to estimate the *ExpertRank* of each user. Liu et al. proposed to consider more pairwise comparisons among questions, askers, non-best answerers, and best answerers [Liu et al., 2011]. For example, given a question and answering thread, it is likely that the expertise score of the best answerer is higher than the asker's and all other non-best answerers'. These pairwise competitions are used as an input into competition-based models or an SVM model [Liu et al., 2011, Aslay et al., 2013] to generate a ranked list of users based on their predicted expertise scores. Bouguessa et al. provide an in-degree method that computes user authority based on the number of best answers provided [Bouguessa et al., 2008]. Users with top authorities have high probabilities to be selected as best answerers.

The topic-oriented user expertise estimation methods are based on latent topic modeling techniques for the content of the questions. CQARank [Yang et al., 2013] was proposed to take both user topical interests and expertise evaluation into consideration. They are able to find experts with both similar topical preference and matching topical expertise. They assumed that every new question falls into some particular topics, and their model is trained on fine-grained topics, which limits its scalability. And their model did not consider the user from the two role perspective (as an asker and as an answerer) as it derived user expertise from questions and answers simultaneously [Xu et al., 2012, Srba and Bielikova, 2016].

---

[8] This forum is accessible from https://www.java-forums.org/forum.php

GRLM [Zhao et al., 2015a] also failed to view the user from the two role perspective. GRLM was proposed to infer the expertise of users and route questions to cold-start experts (users who have answered very few questions), since Zhao et al. discovered that Quora enjoys great benefits contributed by cold-start users. The latent topic model suffers from the data sparsity problem for inferring user features since there are many missing values in cold-start users. GRLM proposed to make use of the user-to-user graph to tackle the data sparsity problem. If two users follow some common topics (interests), there is an edge in the corresponding user-to-user graph. An edge between two users provides a strong evidence for them to have common interests and preferences. SocialTransfer was proposed to transfer social knowledge of users to solve the data sparseness problem in finding cold-start experts [Zhao et al., 2014]. For example, if a cold-start user $u_1$ in Quora has posted sufficient tweet information in Twitter, SocialTransfer can leverage these tweets information to infer the expertise of $u_1$. Apart from inferring the expertise of users from their tweets, SocialTansfer can transfer the knowledge from neighbors of $u_1$ to $u_1$ for inferring $u_1$'s expertise. Similarity among users can be computed by their corresponding follower/followee information in Twitter. Liu et al. tackled the sparsity problem by integrating topic representations from CQA data with network structure from the viewpoint of knowledge graph embedding [Liu et al., 2017]. All objects including question, users, and tags are connected by some relations (ask, belong to and so on). Knowledge graph embedding methods such as TransR [Lin et al., 2015] can be employed to represent the CQA graph. Zhao et al. proposed a topic-level expert learning framework which simultaneously provides the topic of questions and identifies experts on each topic [Zhao et al., 2013]. Xu et al. represents the dual role of users (asker and answerer) via PLSA-based model [Xu et al., 2012]. DCNN modeled the complex matching relations between questions and answers for answer retrieval by using similarity matrix based architectures [Shen et al., 2015]. Besides topic expertise, another factor involving in question routing is availability. Aardvark, a statistical model for routing questions to potential answerers, can prioritize candidate users who are currently online, who are historically active at the present time-of-day, and who have not been contacted recently with a request to answer a question [Horowitz and Kamvar, 2010]. Each candidate user is assigned a score by a scoring function which is composed of a question-dependent relevance score and a question-independent quality score.

To identify expert users more precisely, Huna et al. proposed to model users expertise with accentuation on the quality of users contributions and the difficulty of questions users have answered [Huna et al., 2016]. A user gains greater reputation for asking difficult and useful questions and for providing useful answers on other difficult questions. Hanrahan et al. used the duration between the time when the question was asked and the time when an answer was marked as the best answer as the measure for question difficulty [Hanrahan et al., 2012]. Yang et al. proposed that harder questions can generate more answers or discussions than easier ones. They called the number of answers provided for a question as debatableness, which is a very important factor for determining the expertise of users in their model[Yang et al., 2014].

Unlike previous approaches, MC [Zhao et al., 2015b] formulated the problem of expert finding as a missing value estimation problem, which can, in turn, be cast into a matrix completion optimization problem, based on the past question-answering activities of users in CQAs. However it only holds latent vectors for every existing user/question IDs. There is no way to make a meaningful recommendation under an unforeseen condition. To address the biased estimator raised by using the absolute votes of users' past question-answering activities in existing models[Yang et al., 2013, Zhao et al., 2015b], the relative quality rank was used to model the performance of users for answering the questions [Fang et al., 2016, Zhao et al., 2016, Zhao et al., 2017]. For example AMRNL [Zhao et al., 2017] exploited the relative number of up-votes in the form of quintuple $(i, j, k, o, p)$, meaning that the $j$-th answer provided by the $k$-th user, obtains more up-votes than the $o$-th answer provided by the $p$-th user for the $i$-th question. The relative quality of question-answer pairs are integrated in their proposed asymmetric multi-faceted ranking network, which can rank the answers to the given question and select the answer with the highest score as the best answer. The questions, answers, and users are encoded into fixed embedding vectors based on the variant recurrent neural networks called long short term memory (LSTM). In HSNL [Fang et al., 2016], the questions, answers and users are modeled to utilize the textual contents and the social relationships simultaneously. Above approaches use the answer information – which is unavailable in the cold question routing problem considered in this paper (finding matching experts before answers are written). A social relation between two users provides a strong evidence for them to have common background [Jiang et al., 2015, Zhao et al., 2015b], hence RMNL [Zhao et al., 2016] was proposed to leverage social relations and triplet constraints to tackle question answering problems in CQAs. A triplet constraint denoted as $(i, j, k)$, means that the $i$-th user obtains more votes than the $k$-th user for answering the $j$-th question. RMNL used users' social network follower/followee information to enhance experts finding ability. However, in Stack Exchange sites, it is not easy for us to find users' social relations, and Zhao et al. [Zhao et al., 2015b, Zhao et al., 2016] reported that only about one-third of the users in Quora have a twitter account. MCR [Dror et al., 2011] considered the question routing as a classification task whether a particular question will be interesting for a user or not. They considered question askers and their corresponding question asking history as a channel, which increased the difficulty of routing new questions posted by new askers who have no asking history. And MCR used 530 hand-crafted features, which is not easy to reproduce. QDEE [Sun et al., 2018] proposed to leverage Expertise Gain Assumption (EGA) to avoid the data spareness problem and built competition graphs from the users' past asking and answering activities. QDEE interprets the hierarchy of corresponding competition graph as the question difficulty and user expertise. The corresponding graph hierarchy is inferred by TureSkill [Herbrich et al., 2007] and Social Agony [Tatti, 2014, Tatti, 2015, Sun et al., 2017]. QDEE relies on textual features (to identify semantically similar questions) as well as estimated question difficulty to generate related context, and subsequently uses this to estimate difficulty level of newly posed questions and routes them to appropriate users.

We summarize the differences between the proposed ColdRoute model with some of these recent efforts in Table 1.

Table 1: Comparison of different methods with ColdRoute (short for CR)

| Attributes | CR | SM | QDEE | MCR | CQARank | GRLM |
|---|---|---|---|---|---|---|
| using question tags (categories) | ✓ | ✓ | ✓ | ✓ | ✓ | ✓ |
| involving of answerers | ✓ | ✗ | ✓ | ✓ | ✓ | ✗ |
| involving of askers | ✓ | ✗ | ✓ | ✓ | ✓ | ✗ |
| absolute/relative up-votes of questions | ✓ | ✗ | ✓ | ✗ | ✓ | ✓ |
| topic-free training | ✓ | ✓ | ✓ | ✓ | ✗ | ✓ |
| two-role perspective | ✓ | ✗ | ✓ | ✓ | ✗ | ✗ |
| routing cold questions (existing askers) | ✓ | ✓ | ✓ | ✓ | ✓ | ✓ |
| routing cold questions (new askers) | ✓ | ✓ | ✓ | ✗ | ✓ | ✓ |

## 3 Problem Statement

Assume we are given four relational sets of data in terms of Questions $\mathcal{Q} = \langle q_1, q_2, \ldots, q_n \rangle$, Askers $\mathcal{A} = \langle a_1, a_2, \ldots, a_m \rangle$, Answerers $\mathcal{U} = \langle u_1, u_2, \ldots, u_k \rangle$, and Question Tags $\mathcal{T} = \langle t_1, t_2, \ldots, t_l \rangle$. For each question $q_i \in \mathcal{Q}$, we have a tuple of the form $\langle Asker_i, Answerers_i, BestAnswerer_i, Tags_i, \text{and } Scores_i \rangle$, where $Asker_i \in \mathcal{A}$, $Answerers_i \subset \mathcal{U}$, $BestAnswerer_i \in \mathcal{U}$, $Tags_i \subset \mathcal{T}$. Each voting $score \in Scores_i$ is an integer, which is calculated based on the difference between $Answerer_i$'s up-votes and down-votes which are assigned to it by users who viewed the question or provided answers for that in the CQA environment. Note that the $BestAnswerer$ for a question may not be specified by $Asker$.

Given the preliminaries (above), in this work, we focus on the problem of routing newly posted questions to matching experts before answers are written (item cold-start). Each quadruple case $\langle q, u, a, t \rangle$, where $q \in \mathcal{Q}$, $u \in \mathcal{U}$, $a \in \mathcal{A}$, $t \subset \mathcal{T}$ has a voting score $y \in \mathbb{R}$, which is equal to the difference between times of up-voting and down-voting. Our goal is to learn a function $f : \langle q, u, a, t \rangle \to \mathbb{R}$. The user $u \in \mathcal{U}$ who achieves the highest value of $f(q, u, a, t)$ will be selected as the best answerer for question $q$. Particularly for a newly posted question $q$ asked by a new asker using tags $t$ and a potential answerer $u$, the prediction function $f$ can treat the new asker as a missing value by $f(q, u, 0, t)$. It is possible that the potential answerer $u$ is a newly registered user who has not provided any answer before in CQAs (user cold-start). In this scenario, the prediction function can be simplified as $f(q, 0, 0, t)$. All new registered users will receive the same predicted voting score for the same target question. More efforts will be spent to make accurate predictions for the user cold-start problem in our future work.

## 4 ColdRoute Design

In this section, we describe the architecture of our framework for routing newly posted questions. Figure 3 shows the overall process of the ColdRoute framework. The key steps of our framework are: 1) Encode all past activities; 2) Use FMs to train our model; 3) Routing newly posted questions to potential answerers, identified by predicting voting scores with using the model trained in step 2.

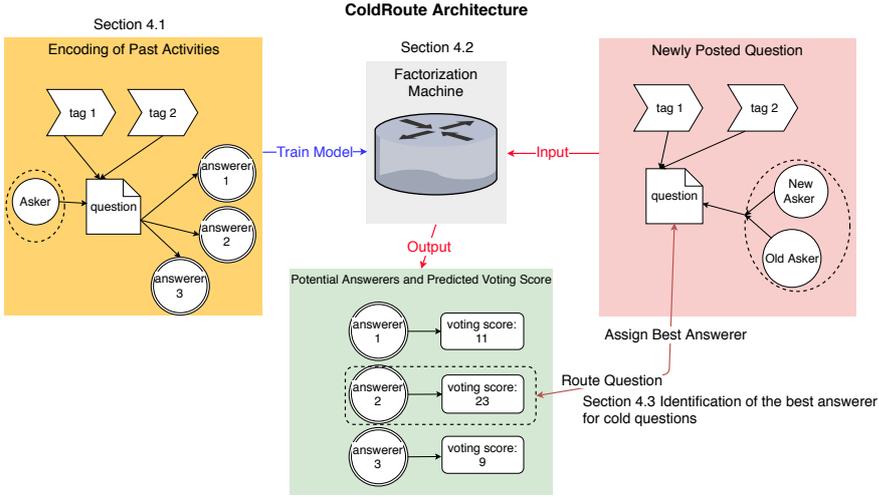

Fig. 3: ColdRoute Architecture: Users' past activities are used to train ColdRoute. Given a newly posted question either it is asked by a new asker or an existing asker, ColdRoute can predict the voting score for each answerer in the candidate set, and then select the user who has the highest voting score as the best answerer to route this cold-start question.

4.1 Encoding of past activities

Table 2 illustrates how we encode all users' past asking and answering activities in CQAs. Our setting can be viewed as a tuple of $(\mathbf{X}, \mathbf{y})$. Let us assume the feature vector matrix $\mathbf{X} \in \mathbb{R}^{n \times p}$, where each row describes an encoding of one quadruple case with $p$ real values and where $\mathbf{y}$ represents the prediction targets (voting scores) of $\mathbf{X}$.

For the $i$-th row $\mathbf{x}^{(i)} \in \mathbb{R}^p$ of $\mathbf{X}$, it represents a quadruple case $\langle q_i, u_i, a_i, t_i \rangle \in \langle \mathcal{Q}, \mathcal{U}, \mathcal{A}, \mathcal{T} \rangle$ as a feature vector $(\mathbf{q}_i, \mathbf{u}_i, \mathbf{a}_i, \mathbf{t}_i)$, where $\mathbf{q}_i$ is the one-hot encoding of $q_i$, $\mathbf{u}_i$ is the one-hot encoding of $u_i$, $\mathbf{a}_i$ is the one-hot encoding of $a_i$, and $\mathbf{t}_i$ encodes all tags in $t_i$. The voting score of $\mathbf{x}^{(i)}$ is $\mathbf{y}^{(i)}$. Suppose the number of unique questions is $|\mathcal{Q}|$, the number of unique answerers is $|\mathcal{U}|$, the number of unique askers is $|\mathcal{A}|$, and the number of unique tags is $|\mathcal{T}|$, we then have $p = |\mathcal{Q}| + |\mathcal{U}| + |\mathcal{A}| + |\mathcal{T}|$. Each feature vector $\mathbf{x}^{(i)}$ has only $(3 + ||\mathbf{t}_i||_1)$ ones. $||\mathbf{t}_i||_1$ represents question $q_i$'s number of tags (number of ones in the vector $\mathbf{t}_i$). Average number of tags per question in our experiments is $2.5^9$. Hence, $\mathbf{X}$ is sparse in our settings.

This design gives us the flexibility to explore the different features' relative importance in cold question routing. A cold question's available information includes: its asker (if previously known), tags, textual descriptions including question head (title) and question body. These features can be iteratively introduced to ColdRoute to explore their relative importance as follows:

– *ColdRoute-A*: explore the importance of question asker by using triples of $\langle \mathcal{Q}, \mathcal{U}, \mathcal{A} \rangle$ on routing cold questions asked by existing askers.

---

[9] Detailed statistics can be seen in Table 3

- *ColdRoute-T*: explore the importance of question tags by using triples of $\langle \mathcal{Q}, \mathcal{U}, \mathcal{T} \rangle$ on routing cold questions either from existing askers or new askers.
- *ColdRoute-TA*: explore the importance of question tags and question asker by using quadruple of $\langle \mathcal{Q}, \mathcal{U}, \mathcal{A}, \mathcal{T} \rangle$ on routing cold questions either from existing askers or new askers.
- *ColdRoute-B*: explore the importance of question body by using triples of $\mathcal{Q}, \mathcal{U}$, and preprocessed question body on routing cold questions either from existing askers or new askers.
- *ColdRoute-H*: explore the importance of question head by using triples of $\mathcal{Q}, \mathcal{U}$, and preprocessed question head on routing cold questions either from existing askers or new askers.
- *ColdRoute-HB*: explore the importance of question textual description by using triples of $\mathcal{Q}, \mathcal{U}$, and preprocessed question head, and preprocessed question body on routing cold questions either from existing askers or new askers.

For data preprocessing of question body and head, we tokenize textual description and discard all code snippets and URLs (if applicable). Then we remove the stop words and HTML tags in the textual description. After stemming, each left term (word) represents a non-zero value in the corresponding feature vector $\mathbf{x}$.

4.2 Factorization Machines

Feature vector $\boldsymbol{X}$ is very sparse since each row of $\boldsymbol{X}$ has a limited number of ones. It is worth mentioning that many traditional machine learning algorithms are not suitable for sparse features. Deep neural network has been applied to many areas successfully recently especially in vision community. However, McMahan et al. discovered that deep neural network does not give a benefit in ad click prediction [McMahan et al., 2013]. The source of difference between the negative results of ad click prediction and the promising results from the vision community lies in the differences in feature distribution. In our task and the ad click prediction task, input features are sparse, while in vision tasks input features are commonly dense. Rendle proposed FMs to handle sparse problems caused by one-hot encoding of user IDs and item IDs in recommender systems [Rendle, 2012, Rendle, 2010].

The reason of FMs being able to handle sparse settings is that FMs can model all nested interactions up to order $d$ between the $p$ variables in $\mathbf{x}$ using factorized interaction parameters [Rendle, 2010, Rendle, 2012]. Consider a 2-way FM ($d = 2$) as an example:

$$\hat{\mathbf{y}}(\mathbf{x}) = w_0 + \sum_{i=1}^{p} w_i x_i + \sum_{i=1}^{p} \sum_{j=i+1}^{p} x_i x_j <\mathbf{v}_i, \mathbf{v}_j> \quad (1)$$

where the model parameters that have to be estimated are:

$$w_0 \in \mathbb{R}, \mathbf{w} \in \mathbb{R}^p, \mathbf{V} \in \mathbb{R}^{p \times k} \quad (2)$$

And $<\cdot, \cdot>$ is the dot product of two vectors of size $k$:

$$<\mathbf{v}_i, \mathbf{v}_j> = \sum_{f=1}^{k} v_{i,f} v_{j,f} \quad (3)$$

where a row $\mathbf{v}_i \in \mathbf{V}$ describes the $i$-th variable with $k \in \mathbb{N}_0^+$ factors. $k$ represents the dimensionality of the factorization.

Table 2: Illustration of FM, the main component in our ColdRoute. Each row represents a feature vector $\mathbf{x}^{(i)}$ and its corresponding target (voting score) $y^{(i)}$. The first 4 columns (orange) represent one-hot encoding of questions (ids); the next 4 (yellow) represent one-hot encoding of answerers (ids); The next 4 columns (blue) hold the one-hot encoding of corresponding askers (ids); The last 4 columns (green) are indicator variables for question tags.

|  | Feature Vector **X** | | | | | | | | | | | | | | | | | Target **y** |
|---|---|---|---|---|---|---|---|---|---|---|---|---|---|---|---|---|---|---|
| $\mathbf{x}^{(1)}$ | 0 | 0 | 1 | $\cdots$ | 1 | 0 | 0 | $\cdots$ | 1 | 0 | 0 | $\cdots$ | 0 | 1 | 0 | $\cdots$ | 4 | $y^{(1)}$ |
| $\mathbf{x}^{(2)}$ | 0 | 0 | 1 | $\cdots$ | 0 | 1 | 0 | $\cdots$ | 1 | 0 | 0 | $\cdots$ | 0 | 1 | 0 | $\cdots$ | 3 | $y^{(2)}$ |
| $\mathbf{x}^{(3)}$ | 0 | 0 | 1 | $\cdots$ | 0 | 0 | 1 | $\cdots$ | 1 | 0 | 0 | $\cdots$ | 0 | 1 | 0 | $\cdots$ | 2 | $y^{(3)}$ |
| $\mathbf{x}^{(4)}$ | 0 | 1 | 0 | $\cdots$ | 0 | 0 | 1 | $\cdots$ | 0 | 0 | 1 | $\cdots$ | 0 | 0 | 1 | $\cdots$ | 5 | $y^{(4)}$ |
| $\mathbf{x}^{(5)}$ | 0 | 1 | 0 | $\cdots$ | 0 | 1 | 0 | $\cdots$ | 0 | 0 | 1 | $\cdots$ | 0 | 0 | 1 | $\cdots$ | 6 | $y^{(5)}$ |
| $\mathbf{x}^{(6)}$ | 1 | 0 | 0 | $\cdots$ | 1 | 0 | 0 | $\cdots$ | 0 | 1 | 0 | $\cdots$ | 1 | 1 | 1 | $\cdots$ | 2 | $y^{(6)}$ |
| $\mathbf{x}^{(7)}$ | 1 | 0 | 0 | $\cdots$ | 0 | 0 | 1 | $\cdots$ | 0 | 1 | 0 | $\cdots$ | 1 | 1 | 1 | $\cdots$ | 4 | $y^{(7)}$ |
|  | $q_1$ | $q_2$ | $q_3$ | $\cdots$ | $u_1$ | $u_2$ | $u_3$ | $\cdots$ | $a_1$ | $a_2$ | $a_3$ | $\cdots$ | $t_1$ | $t_2$ | $t_3$ | $\cdots$ | | |
|  | Question | | | | Answerer | | | | Asker | | | | Question Tags | | | | | |

Above 2-way FM can capture all single and pairwise interactions between variables. And the pairwise interactions can be reformulated:

$$\sum_{i=1}^{p}\sum_{j=i+1}^{p} x_i x_j <\mathbf{v}_i, \mathbf{v}_j>$$
$$= \frac{1}{2}\sum_{i=1}^{p}\sum_{j=1}^{p} x_i x_j <\mathbf{v}_i, \mathbf{v}_j> - \frac{1}{2}\sum_{i=1}^{p} x_i x_i <\mathbf{v}_i, \mathbf{v}_i> \quad (4)$$
$$= \frac{1}{2}\sum_{f=1}^{k}\left(\left(\sum_{i=1}^{p} v_{i,f} x_i\right)^2 - \sum_{i=1}^{p} x_i^2 v_{i,f}^2\right)$$

As we have shown, FMs have a closed model equation that can be computed in linear time. And the model parameters ($w_0$, $\mathbf{w}$ and $\mathbf{V}$) of FMs can be learned efficiently by gradient descent methods as:

$$\frac{\partial}{\partial \theta}\hat{\mathbf{y}}(\mathbf{x}) = \begin{cases} 1, & \text{if } \theta \text{ is } w_0 \\ x_i, & \text{if } \theta \text{ is } w_i \\ x_i \sum_{j=1}^{p} v_{j,f} x_j - v_{i,f} x_i^2, & \text{if } \theta \text{ is } v_{i,f} \end{cases} \quad (5)$$

To capture more interactions, FM can be generalized to a $d$-way FM:

$$\hat{\mathbf{y}}(\mathbf{x}) = w_0 + \sum_{i=1}^{p} w_i x_i +$$
$$\sum_{l=2}^{d}\sum_{j_1=1}^{p}\sum_{j_2=j_1+1}^{p}\cdots\sum_{j_l=j_{l-1}+1}^{p}\left(\prod_{i=1}^{l} x_{j_i}\right)\left(\sum_{f=1}^{k_l}\prod_{i=1}^{l} v_{j_i,f}^{(l)}\right) \quad (6)$$

From equation 1 and 6, we can observe that FMs break the independence of the interaction parameters by factorizing them [Rendle, 2010, Rendle, 2012]. Since the data for one interaction also helps to estimate the parameters for related interactions, FMs can work well in sparse settings. The example in Section 4.3 can make this idea more clear.

4.3 Identification of the best answerer for cold questions

As it is shown in Figure 3, there are two steps in identification of the best answerer for cold questions:

- 1. given a cold question $q$ and a set of potential answerers $C_q$, predict each candidate $u$'s voting score for $q$, where $u \in C_q$
- 2. select the user who achieves the highest voting score as the best answerer for $q$

In this section, we use an example to show why other regression models such as linear and polynomial support vector machines (SVMs) fail in step 1 with sparse settings. Suppose we want to find the best answerer for newly posted question $q_4$ asked by a new asker $a_4$ with tags $t = \{t_1, t_2, t_3\}$, the first step is to use a regression model to predict each candidate answerer $u$'s voting score for $q_4$. The simplest regression model is the linear regression model (linear SVM). Given an input feature vector $\mathbf{x}$, linear SVM can predict $\mathbf{x}$'s output as:

$$\hat{\mathbf{y}}(\mathbf{x}) = w_0 + \sum_{i=1}^{p} w_i x_i, \quad w_0 \in \mathbb{R}, \quad \mathbf{w} \in \mathbb{R}^p \tag{7}$$

It is worth mentioning that linear SVM is a special case of factorization machine (set degree $d = 1$ in equation 6). Suppose we want to predict $u_3$'s voting score for $q_4$, and the corresponding input feature vector is represented as $\mathbf{x}^{(8)}$. The linear SVM model (eq. 7) will predict $\mathbf{x}^{(8)}$ as:

$$\hat{\mathbf{y}}(\mathbf{x}^{(8)})_{lr} = w_0 + w_{q_4} + w_{u_3} + w_{a_4} + \sum_{i=1}^{3} w_{t_i} \tag{8}$$

where interactions among variables (question, asker, answerer, and tags) are missing in comparing with FMs.

The polynomial kernel allows the SVMs to model higher interactions between variables [Rendle, 2010]. For example, the prediction model of polynomial SVMs with $d = 2$ can be written as:

$$\hat{\mathbf{y}}(\mathbf{x}) = w_0 + \sqrt{2} \sum_{i=1}^{p} w_i x_i + \sum_{i=1}^{p} w_{i,i}^{(2)} x_i^2 + \sqrt{2} \sum_{i=1}^{p} \sum_{j=i+1}^{p} w_{i,j}^{(2)} x_i x_j \tag{9}$$

where the model parameters are: $w_0 \in \mathbb{R}$, $\mathbf{w} \in \mathbb{R}^p$, and $\mathbf{W}^{(2)} \in \mathbb{R}^{p \times p}$. Hence, the polynomial SVM model (eq. 9) can predict $\mathbf{x}^{(8)}$ as:

$$\hat{y}(\mathbf{x}^{(8)})_{svr} = w_0 + \sqrt{2}(w_{q_4} + w_{u_3} + w_{a_4} + \sum_{i=1}^{3} w_{t_i})$$
$$+ (w_{q_4,q_4}^{(2)} + w_{u_3,u_3}^{(2)} + w_{a_4,a_4}^{(2)} + \sum_{i=1}^{3} w_{t_i,t_i}^{(2)}) + \sqrt{2}(\sum_{i \in \mathbf{S}} \sum_{j \in \mathbf{S}, i<j} w_{i,j}^{(2)}) \quad (10)$$

where $\mathbf{S} = \{q_4, u_3, a_4, t_1, t_2, t_3\}$. Since $w_{q_4}$ and $w_{q_4,q_4}^{(2)}$ express the same meaning, $\hat{y}(\mathbf{x}^{(8)})_{svr}$ is the same as the linear case $\hat{y}(\mathbf{x}^{(8)})_{lr}$ but with an additional interactions represented as $\sum_{i \in \mathbf{S}} \sum_{j \in \mathbf{S}, i<j} w_{i,j}^{(2)}$. With the polynomial kernel, the SVMs can capture higher-order interactions. However, to have a reliable estimation of the parameter $w_{i,j}^{(2)}$ of a pairwise interaction $(i,j)$, there must be enough cases $\mathbf{x} \in \mathbf{X}$ where $x_i = 1$ and $x_j = 1$. Either $x_i = 0$ or $x_j = 0$ can cause case $\mathbf{x}$ not to be used for estimating the interaction parameter $w_{i,j}^{(2)}$. In our sparse scenarios, there are too few or even no cases for $(i,j)$. Hence, the polynomial SVM can not leverage higher order interactions for predicting test examples and thus cannot provide better estimation than a linear SVM [Rendle, 2010].

Unlike SVMs that all interaction parameters $w_{i,j}^{(2)}$ of SVMs are completely independent, FMs can estimate interactions in sparse settings well because they break the independence of interaction parameters by factorizing them [Rendle, 2010, Rendle, 2012]. The factorized interactions can make FMs model all possible interactions between values in the feature vector $\mathbf{x}$ even under high sparsity. Especially, it is possible to generalize to unobserved interactions. For example, $< \mathbf{v}_{t_1}, \mathbf{v}_{t_2} >$ and $< \mathbf{v}_{t_1}, \mathbf{v}_{t_3} >$ depend on each other as they overlap and share the common parameters $\mathbf{v}_{t_1}$. The data for one interaction $< \mathbf{v}_{t_1}, \mathbf{v}_{t_2} >$ can help to estimate the parameters for related interactions such as $< \mathbf{v}_{t_1}, \mathbf{v}_{t_3} >$.

Suppose we use a 2-way FM to estimate the voting score for $\mathbf{x}^{(8)}$, the first part for estimation is the dot product between $\mathbf{w}$ and $\mathbf{x}^{(8)}$, which is equivalent with linear SVM (eq. 7 and eq. 8). The dominant part for estimation is interactions among $q_4$, $a_4$, $u_3$, $t_1$, $t_2$, and $t_3$. In this example, these interactions can be represented as summation of dot products $\sum_{i \in \mathbf{S}} \sum_{j \in \mathbf{S}, i<j} \langle \mathbf{v}_i, \mathbf{v}_j \rangle$, where $\mathbf{S} = \{q_4, u_3, a_4, t_1, t_2, t_3\}$.

After we predict the voting score for each candidate answerer, we can select the user who achieves the highest voting score as the best answerer for the newly posted question. For example in Figure 3, candidate answerer 2 is identified as the best answerer for the newly posted question.

## 5 Experimental Evaluation

In this section, we evaluate the performance of ColdRoute. First, we consider the performance on resolved questions and compare it with well known techniques. We then compare the results with newly posted questions asked by existing askers and new askers separately. We begin by describing the experimental settings such as datasets, and measures-of-interest.

Table 3: Statistics of Stack Exchange Sites (Ask., Ser. are short for AskUbuntu and Serverfault respectively)

|  | Apple | Ask. | Gaming | Physics | Scifi | Ser. | Tex | Unix |
|---|---|---|---|---|---|---|---|---|
| # Questions | 80,466 | 257,173 | 75,696 | 93,529 | 38,026 | 238,764 | 129,182 | 111,505 |
| # Answers | 119,878 | 337,198 | 130,294 | 137,258 | 78,652 | 398,470 | 169,354 | 171,016 |
| # Unique Users | 65,851 | 189,955 | 51,192 | 41,115 | 26,673 | 130,951 | 48,049 | 65,279 |
| # Questions having Best Answers | 29,765 | 85,843 | 45,798 | 38,094 | 21,740 | 117,275 | 76,862 | 53,856 |
| # Unique Tags | 1,048 | 3,020 | 4,437 | 876 | 2,349 | 3,514 | 1,525 | 2,438 |
| Avg # Tags per Question | 2.824 | 2.6982 | 1.2823 | 2.9634 | 2.1967 | 2.882 | 2.2752 | 2.7868 |
| # Askers | 40,206 | 137,171 | 25,153 | 31,415 | 12,413 | 93,739 | 42,819 | 45,773 |
| # Asker (asked only 1 question) (%) | 76.74% | 75.88% | 74.23% | 63.26% | 74.71% | 64.04% | 62.55% | 68.48% |
| Avg # Questions per Asker | 1.9758 | 1.8557 | 2.9689 | 2.8849 | 3.0031 | 2.4411 | 2.9851 | 2.4022 |

5.1 Experimental Settings

The first step is to describe the Stack Exchange sites which we use for evaluation of our ColdRoute. We select 8 large and popular sites from the most recent data dump of Stack Exchange[10]. More details about the Stack Exchange sites can be found in the Table 3.

5.2 Evaluation Criteria

Our task is to select the user who achieves the highest voting score as the best answerer for a newly posted question. Given the testing question set $\mathcal{Q}$, the predicted ranking of all the answerers for question $q$ is $R^q$. We evaluate the performance of our proposed methods based on several popular evaluation criteria for the problem of expert finding and question routing in CQAs, i.e. Mean Reciprocal Rank (MRR) [Liu et al., 2011, Zhu et al., 2014, Zhu et al., 2011], Precision@$k$ [Zhu et al., 2011, Zhu et al., 2014, Guo et al., 2008, Zhao et al., 2017, Zhao et al., 2016, Fang et al., 2016], and Accuracy [Xu et al., 2012, Zhou et al., 2012a, Zhao et al., 2017, Zhao et al., 2016, Fang et al., 2016].

**MRR.** The MRR measure is given by

$$MRR = \frac{1}{|Q|} \sum_{q \in Q} \frac{1}{r^q_{best}} \quad (11)$$

where $r^q_{best}$ is the position of question $q$'s best answerer in the predicted ranking list. It is worth mentioning that MRR is equivalent to Mean Average Precision (MAP) since the number of correct elements (the best answerer) in the predicted ranking list is just 1.

**Precision@$k$.** The $Precision@k$ is applied to measure the average number of times that the best answerer is ranked on top-$k$ by a certain algorithm.

$$Precision@k = \frac{\{q \in Q | r^q_{best} <= k\}}{|Q|} \quad (12)$$

**Accuracy.** The Accuracy is used to measure the ranking quality of the best answerer, given by

$$Accuracy = \frac{1}{|Q|} \sum_{q \in Q} \frac{|R^q| - r^q_{best}}{|R^q| - 1} \qquad (13)$$

---

[10] We used the data dump which is released on June 12, 2017 and is available online at https://archive.org/details/stackexchange

Where $Accuracy = 1$ (best) means that the best answerer returned by an algorithm always ranks on top while $Accuracy = 0$ means the opposite.

5.3 Performance Comparisons

We compare ColdRoute with several state-of-the-art methods for the problem of expert finding and question routing in CQAs as follows:

- **AuthorityRank (AR)** [Bouguessa et al., 2008] computes the user authority based on the number of best answers provided (AR-ba). AR-a is a modified version to compute the user authority based on the number of answers provided. Given a question $q$, its candidate answerers are ranked based on their authority.
- **BoW** is an answer ranking algorithm based on the bag-of-words representations of both questions and answers (or answerers for the task of routing newly posted questions) for computing the matching score. It has been shown successful in many question answering applications [Figueroa and Neumann, 2013, Zhou et al., 2012b, Zhou et al., 2013]. BoW used in our paper is implemented by scikit-learn [11].
- **Doc2Vec** [Le and Mikolov, 2014, Dong et al., 2015] encodes both questions and answers (or answerers for the task of routing newly posted questions) into a low-dimensional continuous feature space based on the distributed bag-of-words representation for computing the relevant score. The Doc2Vec used in our paper is implemented by gensim [12]. The dimension of the feature vector is tuned to set as 80.
- **LDA** [Guo et al., 2008, Ji et al., 2012] learns latent topics in the content of questions and answers (or answerers for the task of newly posted questions) as well as latent interests of users in CQA sites via LDA-based model. The LDA model used in our paper is implemented by gensim [13]. The number of topics is tuned to set as 100.
- **MC** [Zhao et al., 2015b] is a graph regularized matrix completion model for learning user model from the viewpoint of missing value estimation and providing answer ranking based on the answerers' expertise. It is worth noticing that we don't have the social relation of users in Stack Exchange sites, hence the objective function in our experiments for estimating the missing value is only based on the past question-answering activities of users in CQAs. Moreover, simply by using askers and answerers in the input feature vector, FMs are similar to MC. The MC code used in our paper is from Libpmf[14] [Yu et al., 2012]. The rank is tuned to set as 30.
- **MLP** is a multi-layer perceptron based regressor for heterogeneous CQA network $G$. $G$ is built based on interactions between askers and questions, questions and answers, and answers and answerers. Directions of edges in $G$ are from askers to questions, questions to answers, lower up-votes answers to higher up-votes answers (for the same questions), and answers to answerers. Node2Vec [Grover and et al., 2016] is applied to learn embeddings for question nodes and answerer nodes in $G$ [15]. Given a target question's embedding, MLP can predict its best answerer's embedding. Then MLP searches the candidate answerers and routes the target question to the user who has the most similar embedding with the prediction. MLP is built based on Keras [16]. It has two hidden layers. Each hidden layer has 256 units, which uses sigmoid as the activation function. It is worth mentioning that MLP only uses users' past activities in CQAs without leveraging any textual information.

---

[11] http://scikit-learn.org/stable/modules/feature_extraction.html
[12] https://radimrehurek.com/gensim/models/doc2vec.html
[13] https://radimrehurek.com/gensim/models/ldamodel.html
[14] https://www.cs.utexas.edu/~rofuyu/libpmf/

- **CQARank** [Yang et al., 2013] jointly models Q&A textual content with votes and tags using a probabilistic generative model, and then leverages link analysis in their constructed Q&A graph $G$ to enforce user topical and expertise learning. Users with high topical interests and expertise will be recommended for newly posted questions. The direction of edges in $G$ is from the asker to the answerer. The underlying assumption is that askers have lower expertise than corresponding answerers. However, the expertise of the asker is not assumed to be lower than the expertise score of a non-best answerer, since such a user may just happened to see the question and responded that, rather than knowing the answer well [Wang et al., 2014a]. Take category Python in Stack Overflow for example, it is common to have answers like "method $x$ provided by user $a$ works for Python 2.7, but I have trouble in running it with Python 3.0". These kinds of answers do not show corresponding answerers' expertise are higher than the asker's expertise. The generated noisy edges in CQARank's Q&A graph can undermine CQARank's performance on estimating user expertise.
- **Other Regressors**: To demonstrate the advantages of FMs in sparse settings, we have compared our ColdRoute with several regression models with using the same feature set as the input. We used two types of SVM based regressors implemented by scikit-learn. One is a Epsilon-Support Vector Regression (SVM)[17] with the polynomial kernel (degree is set as 2). Another is the LinearSVR[18] with the kernel type as a linear function. A neural network based regressor (NN) which is implemented based on Keras has also been compared with ColdRoute. NN is a feedforward neural network with three 3 hidden layers containing 512, 256, and 128 units respectively. The activation function is sigmoid. Other parameters such as loss is set as "mean_squared_error", and optimizer is set as "adam".

It is worth noticing that **AuthorityRank**, **MC**, and **MLP** cannot handle the cold-start questions, since newly posted questions have no information of answers, and cannot infer their embedding and latent representations in **MLP** and **MC** respectively. **AuthorityRank** cannot make personalized cold questions routing. We only report their performance on resolved questions. Since Srba et al.[Srba and Bielikova, 2016] and Dong et al. [Dong et al., 2015] have shown the power of **BoW**, **LDA**, and **Doc2Vec** on question routing (semantic matching be-tween representations of questions and potential answerers) recently, and Yang et al. [Yang et al., 2013] has demonstrated the effectiveness of **CQARank** on recommending expert users for newly posted questions in Stack Overflow, we consider these 4 methods as strong competition partners of ColdRoute on cold question routing. We have compared *ColdRoute-T* with **SVR**, **LinearSVR** and **NN** by using the same feature set as the input on cold questions to demonstrate the advantages of ColdRoute in sparse settings.

---

[15] Other embedding methods such as DeepWalk [Perozzi et al., 2014], Line [Tang et al., 2015] and SEANO [Liang et al., ] are also workable.

[16] https://keras.io/

[17] http://scikit-learn.org/stable/modules/generated/sklearn.svm.SVR.html

[18] http://scikit-learn.org/stable/modules/generated/sklearn.svm.LinearSVR.html

Table 4: Number of different type of questions for evaluation

| # valid questions for evaluation | Apple | Ask. | Gaming | Physics | Scifi | Ser. | Tex | Unix |
|---|---|---|---|---|---|---|---|---|
| # resolved questions | 1,735 | 3,642 | 1,935 | 1,426 | 2,054 | 6,098 | 1,316 | 2,375 |
| # cold questions posted by existing askers | 234 | 467 | 313 | 196 | 279 | 945 | 175 | 297 |
| # cold questions posted by new askers | 263 | 459 | 160 | 229 | 161 | 600 | 132 | 377 |

5.4 Performance on Resolved Questions

To better evaluate the performance of different models on Stack Exchange sites, questions used for evaluation have to meet two requirements 1) have at least 5 answers, 2) have the best answer. These kinds of questions are represented as $Q_r$. For each question $q \in Q_r$, we predict the voting score for $q$'s best answerer (the information of non-best answeres will be used for training). We then select the user who has the highest voting score as the best answerer for routing and then compute the corresponding *Accuracy*, *Precision*@*k*, and *MRR*. A 5-folds cross validation is conducted to avoid over-fitting. The number of valid resolved questions for evaluation in this part is shown in Table 4.

Based on our experiments, the ranking of different methods' performance on resolved questions is: ColdRoute ≻ MC ≻ AR-ba ≻ AR-a ≻ (MLP ≈ CQARank) ≻ (LDA ≈ BoW) ≻ Doc2Vec. Table 5 shows performance of these methods on different Stack Exchange sites. We can conclude that:

- *ColdRoute* performs the best. The second best model is *MC*. *MC* can be viewed as a mimic of a basic version of FM (only using answerers and questions for feature vectors). By incorporating information of askers and question tags, *ColdRoute* improves upon the routing metrics (Precision@1, Accuracy, MRR) over *MC* by 10.78%, 4.15%, and 5.59% respectively.
- *AR-ba* performs better than *AR-a*. It is easy to understand that a user who answers 50 questions with 40 best answers can provide more trustworthy and correct information than a user who answers 100 questions with 0 best answers.
- As we already mention in the earlier section, semantic matching based models (*LDA*, *BoW*, *Doc2Vec*) perform the worst. Particularly *Doc2Vec* performs much worse than *LDA* and *BoW*. *MLP* performs better than semantic matching models, which shows that interaction graph based features can provide useful information for routing questions.

Table 5: Performance on resolved questions in 8 different Stack Exchange sites

| | | Apple | Ask. | Gaming | Physics | Scifi | Ser. | Tex | Unix |
|---|---|---|---|---|---|---|---|---|---|
| MRR | AR-a | 0.5682 | 0.5609 | 0.6557 | 0.5475 | 0.6382 | 0.4825 | 0.5151 | 0.5396 |
| | AR-ba | 0.6200 | 0.6121 | 0.6942 | 0.5889 | 0.6659 | 0.5036 | 0.5321 | 0.5715 |
| | MLP | 0.5749 | 0.5444 | 0.5237 | 0.5683 | 0.5694 | 0.5564 | 0.5785 | 0.5790 |
| | BOW | 0.4593 | 0.4760 | 0.4823 | 0.4584 | 0.4623 | 0.4534 | 0.4801 | 0.4776 |
| | Doc2Vec | 0.3452 | 0.3556 | 0.3071 | 0.3233 | 0.2905 | 0.3557 | 0.3806 | 0.3631 |
| | LDA | 0.4605 | 0.4640 | 0.5143 | 0.4805 | 0.4886 | 0.4642 | 0.4586 | 0.4750 |
| | CQARank | 0.4743 | 0.5667 | 0.5336 | 0.6124 | 0.4951 | 0.4657 | 0.5223 | 0.6237 |
| | MC | 0.6898 | 0.7104 | 0.7653 | 0.7269 | 0.7921 | 0.6807 | 0.6741 | 0.7164 |
| | ColdRoute | 0.7316 | 0.7437 | 0.8051 | 0.7685 | 0.8113 | 0.7366 | 0.7294 | 0.7466 |
| Precision@1 | AR-a | 0.3378 | 0.3383 | 0.4517 | 0.3135 | 0.4241 | 0.2514 | 0.2789 | 0.3015 |
| | AR-ba | 0.3810 | 0.3885 | 0.4941 | 0.3612 | 0.4494 | 0.2639 | 0.2948 | 0.3309 |
| | MLP | 0.4012 | 0.3688 | 0.3581 | 0.3955 | 0.4270 | 0.3693 | 0.4027 | 0.4118 |
| | BOW | 0.2444 | 0.2471 | 0.2382 | 0.2195 | 0.2235 | 0.2160 | 0.2394 | 0.2387 |
| | Doc2Vec | 0.1256 | 0.1255 | 0.0894 | 0.0947 | 0.0764 | 0.1292 | 0.1489 | 0.1335 |
| | LDA | 0.2282 | 0.2276 | 0.2853 | 0.2370 | 0.2537 | 0.2297 | 0.2097 | 0.2328 |
| | CQARank | 0.2605 | 0.3247 | 0.2781 | 0.3314 | 0.2474 | 0.3384 | 0.2792 | 0.3787 |
| | MC | 0.5101 | 0.5439 | 0.6109 | 0.5589 | 0.6509 | 0.5090 | 0.4856 | 0.5402 |
| | ColdRoute | 0.5671 | 0.5851 | 0.6724 | 0.6206 | 0.6855 | 0.5845 | 0.5623 | 0.5836 |
| Accuracy | AR-a | 0.7064 | 0.6963 | 0.7788 | 0.6728 | 0.7818 | 0.6059 | 0.6274 | 0.6655 |
| | AR-ba | 0.7771 | 0.7628 | 0.8215 | 0.7227 | 0.8174 | 0.6466 | 0.6519 | 0.7072 |
| | MLP | 0.6259 | 0.6187 | 0.5671 | 0.6369 | 0.6213 | 0.6505 | 0.6448 | 0.6346 |
| | BOW | 0.6055 | 0.6069 | 0.6145 | 0.5698 | 0.6000 | 0.5742 | 0.5849 | 0.4776 |
| | Doc2Vec | 0.4112 | 0.4155 | 0.3355 | 0.3623 | 0.3302 | 0.4233 | 0.4377 | 0.3631 |
| | LDA | 0.5754 | 0.5860 | 0.6343 | 0.5992 | 0.6280 | 0.5855 | 0.5669 | 0.4750 |
| | CQARank | 0.5462 | 0.6799 | 0.5896 | 0.6384 | 0.6332 | 0.69 | 0.5901 | 0.6681 |
| | MC | 0.7944 | 0.8024 | 0.8576 | 0.8155 | 0.8832 | 0.7712 | 0.7684 | 0.8095 |
| | ColdRoute | 0.8339 | 0.8351 | 0.8848 | 0.8483 | 0.8939 | 0.8205 | 0.8174 | 0.8336 |

5.5 Performance on Cold Questions

Newly posted questions can fall into two categories: asked by existing askers, and asked by new registered askers. Existing askers have asked questions before, while new registered askers are unknown in Stack Exchange sites. We tested ColdRoute on these two different types of questions separately.

5.5.1 **Performance on New Questions Posted by Existing Askers**

We use following procedures to select new questions posted by existing askers in Stack Exchange sites for evaluation:

- filter askers who have asked at least 2 questions as $A_2$
- for each asker $a \in A_2$, filter the most recent asked question $q_a$ that satisfies the conditions that $q_a$ has more than 5 answers and $a$ has specified the best answer for $q_a$, and put $q_a$ into set $Q_{ne}$[19]
- all other questions are represented as $Q_o$

Above procedures can guarantee that $Q_o \cap Q_{ne} = \emptyset$ and $\mathcal{A}_{Q_{ne}} \subset \mathcal{A}_{Q_o}$, where $\mathcal{A}_{Q_{ne}}$ ($\mathcal{A}_{Q_o}$) represents the set of askers who have asked questions $Q_{ne}$ ($Q_o$). Quadruples and their corresponding voting score pairs ($\langle \mathcal{Q}_{Q_o}, \mathcal{U}_{Q_o}, \mathcal{A}_{Q_o}, \mathcal{T}_{Q_o} \rangle, \mathcal{Y}_{Q_o}$) are used for training models (ColdRoute and other comparison partners). Quadruples $\langle \mathcal{Q}_{Q_{ne}}, \mathcal{U}_{Q_{ne}}, \mathcal{A}_{Q_{ne}}, \mathcal{T}_{Q_{ne}} \rangle$ and $Q_{ne}$'s corresponding best answerers are used to compute *Accuracy*, *MRR* and *Precision@k*. Number of valid cold questions selected for evaluation by above procedure is shown in Table 4.

---

[19] ne is short for **n**ewly posted questions asked by **e**xisting askers

To better understand information of askers and question tags' role in cold routing, we provide two variants of ColdRoute: *ColdRoute-A* and *ColdRoute-T*. Comparisons between our ColdRoute (and its variants) and other state-of-the-art models can be viewed at Table 6 and Figure 4. We can observe that:

- *ColdRoute-T*, rather than *ColdRoute-TA*, performs the best over all Stack Exchange sites (except MRR, Precision@1, and Accuracy on *Physics*, and Precision@1 on *Serverfault*). In Table 3 we can see that 70% of askers have only asked 1 question, and the average number of questions per asker has asked is only 2.5. It indicates that adding $\mathcal{A}$ in feature vectors increase the data sparsity, and can not provide enough interactions between askers and other variables (questions, answerers, question tags).
- With increasing information of askers, *ColdRoute-TA*, leveraging more interactions between askers and other variables, can become more robust and efficient. To demonstrate this, we divide Stack Exchange sites into 2 categories: a) *Apple*, *AskUbuntu*, and *Gaming*, b) *Serverfault*, *Tex*, and *Unix* based on the percentage of askers who have asked only 1 question. 75.62% of askers having asked only 1 question among category a, and *ColdRoute-T* improves over *ColdRoute-TA* upon MRR, Precision@1, Precision@3, and Accuracy by 9.87%, 18.03%, 13.66%, and 10.74% respectively. while 65.02% of askers having asked only 1 question among category b, *ColdRoute-T* improves over *ColdRoute-TA* upon MRR, Precision@1, Precision@3, and Accuracy by 2.41%, 3.32%, 3.96%, and 3.04% respectively. As the Stack Exchange sites continue to grow and askers post more and more questions, it is reasonable to believe *ColdRoute-TA* will become more robust and efficient.
- We can observe similar performance patterns of ColdRoute and its variants on 7 Stack Exchange sites, except site *Physics*. For example, in site *Physics*, *ColdRoute-A* performs better than *ColdRoute-T* on MRR and Precision@1. In Table 3 we can observe that site *Physics* has the least number of unique tags, and the proportion of the number of unique tags [20] is only 0.646%. Above settings limit the performance of *ColdRoute-T* and *ColdRoute-TA*.
- Question tags play a more important role than information of askers. Averagely, *ColdRoute-T* improves over *ColdRoute-A* upon MRR, Precision@1, Precision@3, and Accuracy by 6.53%, 11.1%, 10.58%, and 7.81% respectively.
- ColdRoute models perform better than CQARank on almost all datasets (except Tex). In addition to leveraging noisy edges in CQARanks Q&A graph to estimate user expertise, CQARank fails to consider the user from the two role perspective (as an asker and as an answerer) introduced by Xu et al. [Xu et al., 2012] as it derived user expertise from questions and answers simultaneously [Srba and Bielikova, 2016]. Both can undermine CQARank's performance on routing users for cold question.
- Our ColdRoute models can perform better than semantic matching models (using LDA, BOW, and Doc2Vec to represent questions and answers). The results of CQARank are better than semantic matching models too, which indicates the effectiveness of combining topic feature and link structure to improve question routing.

---

[20] $\frac{|\mathcal{T}|}{|\mathcal{T}|+|\mathcal{Q}|+|\mathcal{U}|}$, where $|\mathcal{T}|+|\mathcal{Q}|+|\mathcal{U}|$ is the length of the feature vector used by *ColdRoute-T*

Table 6: Performance on newly posted questions asked by existing askers in 8 different Stack Exchange sites

|  |  | Apple | Ask. | Gaming | Physics | Scifi | Ser. | Tex | Unix |
|---|---|---|---|---|---|---|---|---|---|
| MRR | BOW | 0.3197 | 0.3423 | 0.2908 | 0.343 | 0.2772 | 0.3701 | 0.3504 | 0.4346 |
|  | Doc2Vec | 0.3481 | 0.3605 | 0.2797 | 0.3226 | 0.2979 | 0.3532 | 0.3878 | 0.4044 |
|  | LDA | 0.3567 | 0.3658 | 0.3388 | 0.3956 | 0.3419 | 0.3990 | 0.3557 | 0.4745 |
|  | Linear/SVM/NN | 0.4271 | 0.3993 | 0.3747 | 0.4354 | 0.4153 | 0.4051 | 0.4112 | 0.4043 |
|  | CQARank | 0.4915 | 0.4652 | 0.4463 | 0.5316 | 0.4628 | 0.4627 | 0.4536 | 0.5258 |
|  | ColdRoute-T | 0.5365 | 0.5257 | 0.6445 | 0.5288 | 0.6462 | 0.4792 | 0.4860 | 0.5434 |
|  | ColdRoute-A | 0.4981 | 0.5025 | 0.5884 | 0.5472 | 0.5778 | 0.4609 | 0.4686 | 0.4756 |
|  | ColdRoute-TA | 0.4698 | 0.5016 | 0.5841 | 0.5432 | 0.6213 | 0.4711 | 0.4753 | 0.5263 |
| Precision@1 | BOW | 0.0855 | 0.1092 | 0.0735 | 0.102 | 0.0466 | 0.1238 | 0.1029 | 0.2626 |
|  | Doc2Vec | 0.1197 | 0.1263 | 0.0415 | 0.0918 | 0.0931 | 0.1238 | 0.1314 | 0.2323 |
|  | LDA | 0.1197 | 0.1221 | 0.0927 | 0.1429 | 0.086 | 0.1556 | 0.1143 | 0.2862 |
|  | Linear/SVM/NN | 0.2051 | 0.1585 | 0.1438 | 0.1837 | 0.1864 | 0.1725 | 0.1771 | 0.1616 |
|  | CQARank | 0.2821 | 0.2377 | 0.2269 | 0.2857 | 0.2330 | 0.2370 | 0.1886 | 0.2997 |
|  | ColdRoute-T | 0.3291 | 0.3255 | 0.4505 | 0.2959 | 0.4695 | 0.2519 | 0.2457 | 0.3232 |
|  | ColdRoute-A | 0.2778 | 0.2998 | 0.3898 | 0.3520 | 0.3799 | 0.2243 | 0.2457 | 0.2559 |
|  | ColdRoute-TA | 0.2521 | 0.3041 | 0.3866 | 0.3265 | 0.4265 | 0.2529 | 0.2343 | 0.3064 |
| Precision@3 | BOW | 0.3547 | 0.3940 | 0.2812 | 0.3776 | 0.2796 | 0.4720 | 0.4286 | 0.4444 |
|  | Doc2Vec | 0.3889 | 0.4133 | 0.2971 | 0.3571 | 0.2760 | 0.3979 | 0.5029 | 0.4175 |
|  | LDA | 0.4103 | 0.4411 | 0.4121 | 0.5204 | 0.4229 | 0.5111 | 0.4114 | 0.5017 |
|  | Linear/SVM/NN | 0.5214 | 0.5139 | 0.4441 | 0.5612 | 0.5090 | 0.5026 | 0.5314 | 0.4949 |
|  | CQARank | 0.5855 | 0.5931 | 0.5144 | 0.6990 | 0.5520 | 0.5704 | 0.6457 | 0.6700 |
|  | ColdRoute-T | 0.6581 | 0.6274 | 0.7796 | 0.7194 | 0.7742 | 0.6074 | 0.6343 | 0.6869 |
|  | ColdRoute-A | 0.6026 | 0.5889 | 0.7157 | 0.6582 | 0.6846 | 0.5799 | 0.5657 | 0.5690 |
|  | ColdRoute-TA | 0.5641 | 0.5717 | 0.6805 | 0.6939 | 0.7599 | 0.5778 | 0.6114 | 0.6667 |
| Accuracy | BOW | 0.3893 | 0.4200 | 0.3200 | 0.4089 | 0.3302 | 0.4711 | 0.4189 | 0.4346 |
|  | Doc2Vec | 0.3076 | 0.4333 | 0.3315 | 0.3641 | 0.3097 | 0.4160 | 0.4897 | 0.4044 |
|  | LDA | 0.4485 | 0.4648 | 0.4409 | 0.4946 | 0.4616 | 0.5068 | 0.4268 | 0.4745 |
|  | Linear/SVM/NN | 0.5307 | 0.5017 | 0.4582 | 0.5283 | 0.5144 | 0.4977 | 0.4992 | 0.4765 |
|  | CQARank | 0.5555 | 0.5571 | 0.4979 | 0.6483 | 0.5693 | 0.5562 | 0.5658 | 0.6134 |
|  | ColdRoute-T | 0.6324 | 0.6054 | 0.7387 | 0.6354 | 0.7369 | 0.5807 | 0.5802 | 0.6404 |
|  | ColdRoute-A | 0.5822 | 0.5814 | 0.6710 | 0.6159 | 0.6690 | 0.5655 | 0.5498 | 0.5422 |
|  | ColdRoute-TA | 0.5573 | 0.5671 | 0.6596 | 0.6381 | 0.7174 | 0.5579 | 0.5727 | 0.6174 |

– With using the same sparse feature set as *ColdRoute-T*, LinearSVR, SVM and NN have the similar performance, which is better than semantic matching based models but worse than CQARank and *ColdRoute-T*, as shown in Table 6 and Figure 4. It is consistent with our analysis in Section 4.3. To save space, we use Linear/SVM/NN to represent the best performance among LinearSVR, SVM and NN in Table 6, Table 7, Figure 4, and Figure 5 which demonstrate the advantages of ColdRoute in sparse settings.

### 5.5.2 Performance on New Questions Posted by New Askers

We conduct the following procedure to select new questions posted by new askers for evaluation:

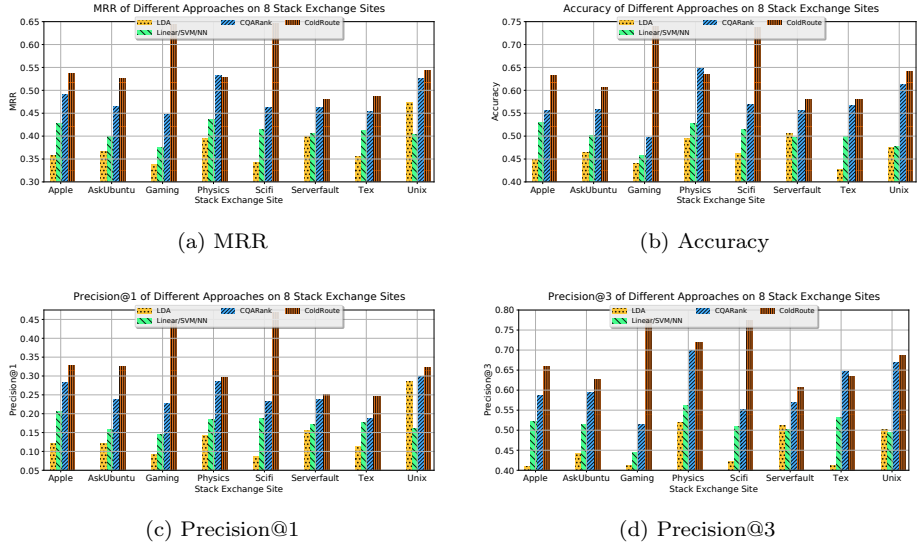

Fig. 4: Performance of ColdRoute-T, different kinds of regressors (with using the same feature set as ColdRoute-T), CQARank and LDA for cold questions asked by existing askers on 8 different Stack Exchange sites

- filter askers who have asked only 1 question as $A_1$
- for each asker $a \in A_1$, filter the question $q_a$ that satisfies the conditions that $q_a$ has more than 5 answers and $a$ has specified the best answer for $q_a$, and put $q_a$ into set $Q_{nn}$[21]
- all other questions are represented as $Q_{o'}$

Above procedure can guarantee that $Q_{o'} \cap Q_{nn} = \emptyset$ and $\mathcal{A}_{Q_{nn}} \cap \mathcal{A}_{Q_{o'}} = \emptyset$, where $\mathcal{A}_{Q_{nn}}$ ($\mathcal{A}_{Q_{o'}}$) represents the set of askers who have asked questions $Q_{nn}$ ($Q_{o'}$). Quadruples and their corresponding voting score pairs ($\langle \mathcal{Q}_{Q_{o'}}, \mathcal{U}_{Q_{o'}}, \mathcal{A}_{Q_{o'}}, \mathcal{T}_{Q_{o'}} \rangle, \mathcal{Y}_{Q_{o'}}$) are used for training models (ColdRotue and other comparison partners). Quadruples $\langle \mathcal{Q}_{Q_{nn}}, \mathcal{U}_{Q_{nn}}, \mathcal{A}_{Q_{nn}}, \mathcal{T}_{Q_{nn}} \rangle$ and $Q_{nn}$'s best answerers are used to compute *Accuracy*, *MRR* and *Precision*@*k*. Number of valid cold questions selected for evaluation by above procedure is shown in Table 4.

Since $\mathcal{A}_{Q_{nn}} \cap \mathcal{A}_{Q_{o'}} = \emptyset$, the asker part in the feature vector used to make predictions are considered as missing values, and **0** is used to represent the feature vector of a new (unseen) asker. Same as Section 5.5.1, *ColdRoute-T*, which use triples $\langle \mathcal{Q}, \mathcal{U}, \mathcal{T} \rangle$ to train and test, is also implemented to make a comparison with *ColdRoute-TA*.

We also leveraged textual descriptions of questions such question head (title) and question body to train ColdRoute. Comparisons between our ColdRoute (and its variants) and state-of-the-art models are shown in the Table 7 and Figure 5. We can observe that:

- *ColdRoute-T* have a comparable performance as *ColdRoute-TA*. As we discussed in Section 5.5.1 and Table 3, 70% of askers have only asked 1 question. It

---
[21] nn is short for **n**ewly posted questions asked by **n**ew askers

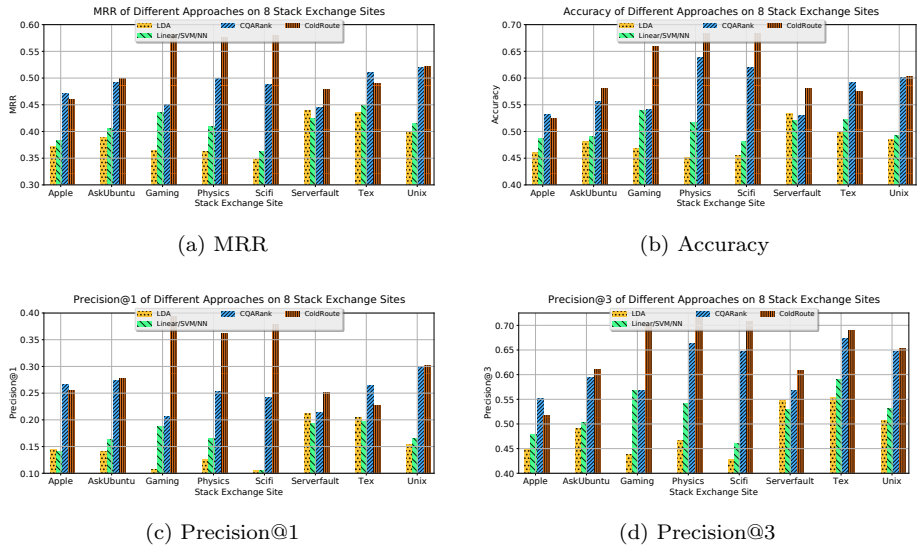

(a) MRR

(b) Accuracy

(c) Precision@1

(d) Precision@3

Fig. 5: Performance of ColdRoute-T, different kinds of regressors (with using the same feature set as ColdRoute-T), CQARank and LDA for cold questions asked by new askers on 8 Stack Exchange sites

- explains that treating unseen askers as missing values and representing them as **0** does not hurt the *ColdRoute-TA* too much.
- *ColdRoute-T* performs better than *ColdRoute-H*. *ColdRoute-H* performs better than *ColdRoute-B*. The question head (title) is the summary of the question (body), and question tags are fine-grained category information of the question. It indicates that ColdRoute favors more general information in terms of cold question routing.
- Our ColdRoute and its variants perform better than semantic matching models (using LDA, BOW, and Doc2Vec to represent questions and answerers).
- Overall, *ColdRoute-T* and *ColdRoute-TA* perform better than Linear/SVM/NN and CQARank, which is consistent with their performance on cold questions asked by existing askers as shown in Section 5.5.1.

## 6 Conclusion

In this paper, we have presented ColdRoute, a framework for tacking cold questions routing in Stack Exchange sites. Specifically, we have used Factorization Machines (FMs) on the one-hot encoding of critical features (question tags and askers) and it can handle cold questions from new or existing askers. By iteratively introducing questions tags and askers, we have observed that question tags play a more important role than information of askers. In Stack Exchange sites, 70% of askers have only asked only 1 question and the average number of questions per asker has asked is only 2.5. Above settings limit information of askers' role in ColdRoute. Generally, a variant of ColdRoute named as *ColdRoute-T*, with using questions, answerers, and question tags, can be deployed to route cold question, either from new (unseen) askers or existing askers in CQAs with sparse askers. With CQAs growing and information of askers becoming dense, *ColdRoute-TA* will be more robust and efficient.

Table 7: Performance on newly posted questions asked by new askers in 8 different Stack Exchange sites

|  |  | Apple | Ask. | Gaming | Physics | Scifi | Ser. | Tex | Unix |
|---|---|---|---|---|---|---|---|---|---|
| MRR | BOW | 0.3515 | 0.3649 | 0.3027 | 0.3287 | 0.2886 | 0.4121 | 0.4129 | 0.4010 |
|  | Doc2Vec | 0.3469 | 0.3630 | 0.3044 | 0.3314 | 0.2789 | 0.3650 | 0.3878 | 0.3660 |
|  | LDA | 0.3713 | 0.3891 | 0.3637 | 0.3626 | 0.3477 | 0.4400 | 0.4351 | 0.4006 |
|  | ColdRoute-HB | 0.3952 | 0.4021 | 0.3840 | 0.3955 | 0.3495 | 0.4116 | 0.4582 | 0.4149 |
|  | ColdRoute-B | 0.4076 | 0.4121 | 0.4034 | 0.4127 | 0.3646 | 0.4165 | 0.4242 | 0.4141 |
|  | ColdRoute-H | 0.4025 | 0.4292 | 0.4198 | 0.4137 | 0.3843 | 0.4194 | 0.4180 | 0.4041 |
|  | Linear/SVM/NN | 0.3839 | 0.4048 | 0.4361 | 0.4102 | 0.3624 | 0.4239 | 0.4490 | 0.4150 |
|  | CQARank | 0.4716 | 0.4913 | 0.4507 | 0.4986 | 0.4873 | 0.4454 | 0.5108 | 0.5205 |
|  | ColdRoute-T | 0.4601 | 0.4988 | 0.5812 | 0.5751 | 0.5793 | 0.4792 | 0.4892 | 0.5214 |
|  | ColdRoute-TA | 0.4698 | 0.4973 | 0.5615 | 0.5689 | 0.5714 | 0.4619 | 0.4910 | 0.5236 |
| Precision@1 | BOW | 0.1293 | 0.1285 | 0.0625 | 0.1004 | 0.0745 | 0.1733 | 0.1591 | 0.1671 |
|  | Doc2Vec | 0.1331 | 0.1285 | 0.0625 | 0.1048 | 0.0621 | 0.1400 | 0.1439 | 0.1247 |
|  | LDA | 0.1445 | 0.1416 | 0.1063 | 0.1266 | 0.1056 | 0.2117 | 0.2045 | 0.1538 |
|  | ColdRoute-HB | 0.1483 | 0.1649 | 0.1438 | 0.1441 | 0.0994 | 0.1778 | 0.2348 | 0.1777 |
|  | ColdRoute-B | 0.1825 | 0.1734 | 0.1813 | 0.1703 | 0.1429 | 0.1850 | 0.1667 | 0.1698 |
|  | ColdRoute-H | 0.1711 | 0.1852 | 0.1875 | 0.1572 | 0.1429 | 0.1883 | 0.1439 | 0.1538 |
|  | Linear/SVM/NN | 0.1407 | 0.1634 | 0.1875 | 0.1659 | 0.1056 | 0.1933 | 0.1970 | 0.1644 |
|  | CQARank | 0.2662 | 0.2745 | 0.2062 | 0.2533 | 0.2422 | 0.2133 | 0.2652 | 0.2997 |
|  | ColdRoute-T | 0.2548 | 0.2767 | 0.3938 | 0.3624 | 0.3789 | 0.2519 | 0.2273 | 0.3024 |
|  | ColdRoute-TA | 0.2471 | 0.2789 | 0.3688 | 0.3537 | 0.3727 | 0.2367 | 0.2424 | 0.3183 |
| Precision@3 | BOW | 0.3840 | 0.4357 | 0.3000 | 0.3799 | 0.2484 | 0.5183 | 0.5530 | 0.5066 |
|  | Doc2Vec | 0.3840 | 0.4096 | 0.3563 | 0.3493 | 0.2547 | 0.4000 | 0.4848 | 0.4297 |
|  | LDA | 0.4487 | 0.4902 | 0.4375 | 0.4672 | 0.4286 | 0.5467 | 0.5530 | 0.5066 |
|  | ColdRoute-HB | 0.5133 | 0.4989 | 0.4625 | 0.5109 | 0.4534 | 0.5090 | 0.5455 | 0.4934 |
|  | ColdRoute-B | 0.4829 | 0.5139 | 0.4563 | 0.5284 | 0.3975 | 0.5133 | 0.5909 | 0.5146 |
|  | ColdRoute-H | 0.4829 | 0.5468 | 0.5063 | 0.5633 | 0.4907 | 0.5050 | 0.5985 | 0.5305 |
|  | Linear/SVM/NN | 0.4791 | 0.5033 | 0.5688 | 0.5415 | 0.4596 | 0.5300 | 0.5909 | 0.5305 |
|  | CQARank | 0.5513 | 0.5948 | 0.5688 | 0.6638 | 0.6460 | 0.5683 | 0.6742 | 0.6472 |
|  | ColdRoute-T | 0.5171 | 0.6100 | 0.7000 | 0.7249 | 0.7081 | 0.6074 | 0.6894 | 0.6525 |
|  | ColdRoute-TA | 0.5589 | 0.6013 | 0.6688 | 0.7205 | 0.6957 | 0.5617 | 0.6439 | 0.6419 |
| Accuracy | BOW | 0.4094 | 0.4323 | 0.3602 | 0.4053 | 0.3199 | 0.5102 | 0.4845 | 0.4667 |
|  | Doc2Vec | 0.4000 | 0.4138 | 0.3779 | 0.3803 | 0.3044 | 0.4233 | 0.4308 | 0.4274 |
|  | LDA | 0.4612 | 0.4815 | 0.4674 | 0.4510 | 0.4542 | 0.5345 | 0.4973 | 0.4859 |
|  | ColdRoute-HB | 0.5104 | 0.5108 | 0.4697 | 0.4948 | 0.4605 | 0.5091 | 0.5084 | 0.4816 |
|  | ColdRoute-B | 0.4851 | 0.5154 | 0.4822 | 0.5150 | 0.4428 | 0.5063 | 0.4970 | 0.4845 |
|  | ColdRoute-H | 0.4909 | 0.5176 | 0.5117 | 0.5220 | 0.4973 | 0.5017 | 0.5003 | 0.4885 |
|  | Linear/SVM/NN | 0.486 | 0.4897 | 0.5394 | 0.5167 | 0.4821 | 0.5203 | 0.5233 | 0.4926 |
|  | CQARank | 0.5316 | 0.5564 | 0.5415 | 0.6389 | 0.6205 | 0.5293 | 0.5925 | 0.6019 |
|  | ColdRoute-T | 0.5247 | 0.5809 | 0.6591 | 0.6838 | 0.6841 | 0.5807 | 0.5747 | 0.6034 |
|  | ColdRoute-TA | 0.5555 | 0.5797 | 0.6327 | 0.6810 | 0.6761 | 0.5460 | 0.5700 | 0.5995 |

As a future work, we plan to test our models on other CQAs with different settings (such as having more dense askers). In order to increase the expertise of the entire community, we plan to address the problem of routing newly posted questions (item cold-start) to newly registered users (user cold-start) in CQAs.

**Acknowledgments** This work is supported by NSF grants CCF-1645599, IIS-1550302, and CNS-1513120, and a grant from the Ohio Supercomputer Center (PAS0166).
**Code and Data**: https://github.com/zhenv5/ColdRoute.